\documentclass{article}

\usepackage{arxiv}

\usepackage[T1]{fontenc}    
\usepackage{hyperref}       
\usepackage{url}            
\usepackage{booktabs}       
\usepackage{amsfonts}       
\usepackage{nicefrac}       
\usepackage{microtype}      
\usepackage{longtable}
\usepackage{booktabs}
\usepackage{xfrac}
\usepackage{amssymb}
\usepackage{times}
\usepackage{latexsym}
\usepackage{tabulary}
\usepackage{algorithm}
\usepackage{algorithmic}
\usepackage{txfonts}
\usepackage{adjustbox}

\title{autoNLP: NLP Feature Recommendations for Text Analytics Applications}

\author{
	Janardan Misra\\
	Accenture Labs\\
	\texttt{janardanmisra@acm.org}
}

\date{}

\begin{document}
	
\maketitle
	
\begin{abstract} 
While designing machine learning based text analytics applications, often, NLP data scientists manually determine which NLP features to use based upon their knowledge and experience with related problems. This results in increased efforts during feature engineering process and renders automated reuse of features across semantically related applications inherently difficult. In this paper, we argue for standardization in feature specification by outlining structure of a language for specifying NLP features and present an approach for their reuse across applications to increase likelihood of identifying optimal features. 
\end{abstract} 


\section{Introduction}
\label{intro}
For an ever increasing spectrum of applications (e.g., medical text analysis, opinion mining, sentiment analysis, social media text analysis, customer intelligence, fraud analytics etc.) mining and analysis of unstructured natural language text data is necessary~\cite{berry2010text,aggarwal2012mining,struhl2015practical}. 

One of key challenge while designing such text analytics (TA) applications is to identify right set of features. For example, for \textit{text classification} problem, different sets of features have been considered in different works (spanning a history of more than twenty years) including `bag of words', `bag of phrases', `bag of n-grams', `WordNet based word generalizations', and `word embeddings'~\cite{tc1,tc2,tc3,tc4,tc5}. Even for recent end-to-end designs using deep neural networks, specification of core features remains manually driven~\cite{DNNforNLPArticle,DNNforNLPBook}. During feature engineering, often data scientists manually determine which features to use based upon their experience and expertise with respect to the underlying application domain as well as state-of-the-art tools and techniques. Different tools (e.g., NLTK~\cite{nltk}, Mallet~\cite{mallet}, Stanford CoreNLP~\cite{stanfordCoreNLP}, Apache OpenNLP~\cite{openNLP}, Apache Lucene~\cite{lucene}, etc.) available to a NLP data scientist for TA application design and development often differ in terms of support towards extraction of features, level of granularity at which feature extraction process is to be specified; and these tools often use different programing vocabularies to specify semantically equivalent features.

Currently, there is no generic method or approach, which can be applied during TA application's design process to define and extract features for any arbitrary application in an automated or semi-automated manner. Even there is no single way to express wide range of NLP features, resulting into increased efforts during feature engineering which has to start new for each data scientist and automated reuse of features across semantically similar or related applications designed by different data scientists is difficult. This also hinders foundational studies on NLP feature engineering including why certain features are more critical than others.  

In this paper, we aim to present an approach towards automating NLP feature engineering. We start with an outline of a language for expressing NLP features abstracting over the feature extraction process, which often implicitly captures intent of the NLP data scientist to extract specific features from given input text. We next discuss a method to enable automated reuse of features across semantically related applications when a corpus of feature specifications for related applications is available. Proposed language and system would help achieving reduction in manual effort towards design and extraction of features, would ensure standardization in feature specification process, and could enable effective reuse of features across similar and/or related applications. 

\section{Life Cycle View}
\begin{figure}[ht]
\begin{center}
\centerline{\includegraphics[width=\textwidth]{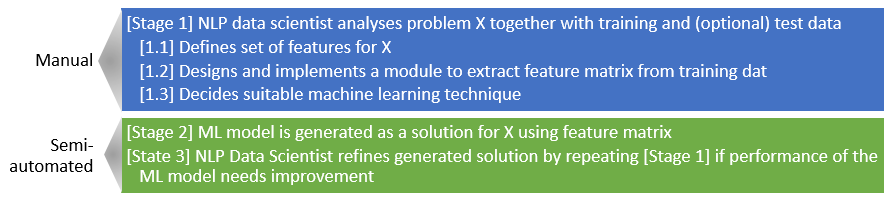}}
\caption{Solution Life Cycle of a (traditional) ML based TA Application}
\label{Fig1}
\end{center}
\end{figure}

Figure~\ref{Fig1} depicts typical design life cycle of a (traditional) ML based solution for the TA applications, which involves steps to manually define relevant features and implement code components to extract those feature from input text corpus during training, validation, testing and actual usage of the application. In traditional ML based solutions, feature interactions also need to be explicitly specified, though this step is largely automated when using deep neural network based solutions~\cite{DNNforNLPBook}. 

As the process of defining features is manual, prior experience and expertize of the designer affects which features to extract and how to extract these features from input text. Current practice lacks standardization and automation in feature definition process, provides partial automation in extraction process, and does not enable automated reuse of features across related application. 

Next, let us consider scenarios when features are specified as elements of a language. Let us refer to this language as \textit{NLP Feature Specification Language} (nlpFSpL) such that a program in nlpFSpL would specify which features should be used by the underlying ML based solution to achieve goals of the target application. Given a corpus of unstructured natural language text data and a specifications in the nlpFSpL, an interpreter can be implemented as feature extraction system (FExSys) to automatically generate feature matrix which can be directly used by underlying ML technique. 

\begin{figure}[ht]
\begin{center}
\centerline{\includegraphics[width=\textwidth]{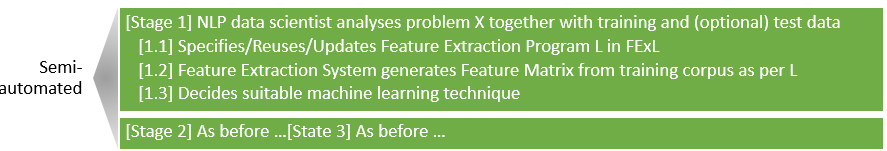}}
\caption{Refined Solution Life Cycle of a ML based TA Applications}
\label{Fig3}
\end{center}
\end{figure}

In contrast to the life-cycle view in the Figure~\ref{Fig1}, this would result into refined solution life cycle of ML based TA applications as depicted in the Figure~\ref{Fig3}.

\section{NLP Feature Specification Language}

\subsection{Meta Elements}
\begin{figure}[ht]
\begin{center}
\centerline{\includegraphics[width=\textwidth]{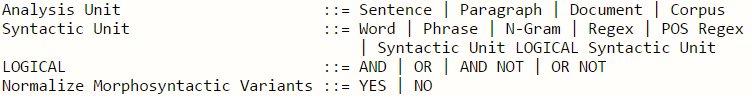}}
\caption{Meta Elements of nlpFSpL}
\label{Fig:ME}
\end{center}
\end{figure}

Figure~\ref{Fig:ME} specifies the meta elements of the nlpFSpL which are used by the FExSys while interpreting other features.  

\begin{figure}[ht]
\begin{center}
\centerline{\includegraphics[width=\textwidth]{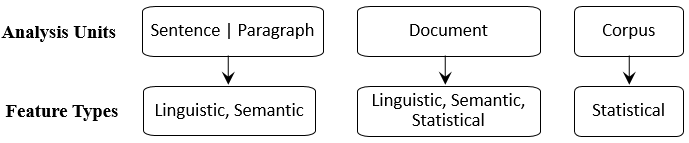}}
\caption{Association between Different Feature Types and Units of Analysis}
\label{Fig5}
\end{center}
\end{figure}

\textit{Analysis Unit} (AU) specifies level at which features have to be extracted. At \textsl{Corpus} level, features are extracted for all the text documents together. At \textsl{Document} level,  features are extracted for each document in corpus separately. At \textit{Para} (paragraph) level Features are extracted for multiple sentences constituting paragraphs together. At \textsl{Sentence} level features to be extracted for each sentence. Figure~\ref{Fig5} depicts classes of features considered in nlpFSpL and their association with different AUs. 

\textit{Syntactic Unit} (SU) specifies unit of linguistic features. It could be a `Word' or a `Phrase', or a `N-gram' or a sequence of words matching specific lexico-syntactic pattern captured as `POS tag pattern' (e.g., Hearst pattern~\cite{hearst}) or a sequence of words matching specific regular expression `Regex' or a combination of these. Option \textit{Regex} is used for special types of terms, e.g., Dates, Numbers, etc. \textsf{\small LOGICAL} is a Boolean logical operator including AND, OR and NOT (in conjunction with other operator). For example, \textsf{\small Phrase AND POS Regex} would specify inclusion of a `Phrase' as SU when its constituents also satisfy 'regex' of `POS tags'. Similarly, \textsf{\small POS Regex OR NOT(Regex)} specifies inclusion of sequence of words as SU if it satisfies `POS tag Pattern' but does not match pattern specified by character `Regex'. Note that SU can be a feature in itself for document and corpus level analysis.

\textit{Normalize Morphosyntactic Variants}: If \textit{YES}, variants of words including stems, lemmas, and fuzzy matches will be identified before analyzing input text for feature exaction and would be treated equivalent.

\subsection{Feature Types}
\subsubsection{Linguistic Features}
\begin{figure*}[ht]
\begin{center}
\centerline{\includegraphics[width=\textwidth]{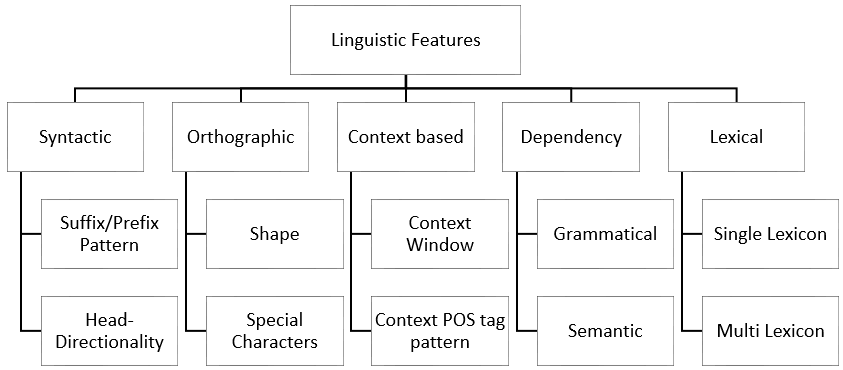}}
\caption{Two level taxonomy of linguistic features}
\label{Fig7}
\end{center}
\end{figure*}

\begin{figure*}[ht]
\begin{center}
\centerline{\includegraphics[width=\columnwidth]{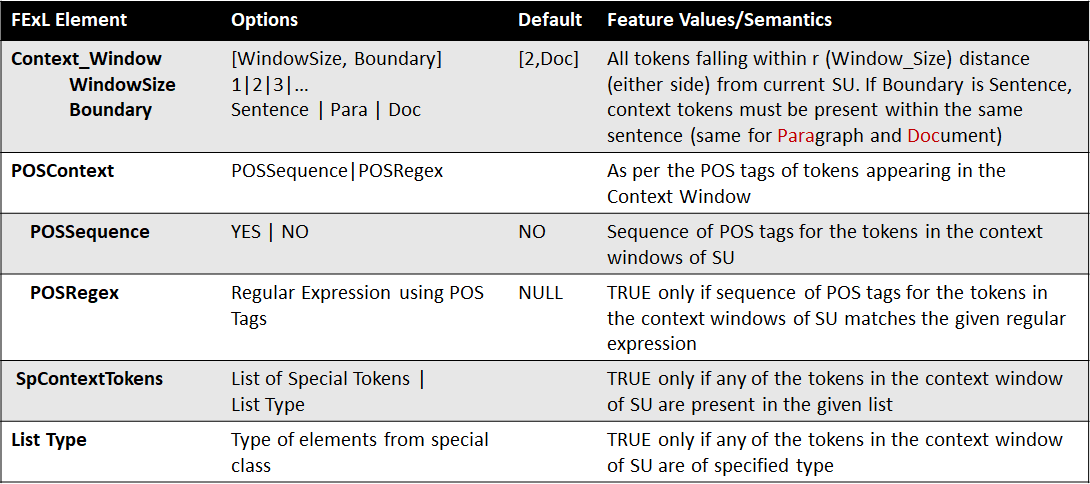}}
\caption{Context based Features}
\label{HD}
\end{center}
\end{figure*}

Figure~\ref{Fig7} depicts two levels of taxonomy for features considered as linguistic. 
To illustrate, let us consider context based features: Table~\ref{HD} gives various options which need to be specified for directing how context for an SU should be extracted. For example, {\sf Context\_Window := [2, Sentence]} will extract all tokens within current sentence, which are present within a distance of 2 on both sides from the current SU. However,  {\sf Context\_Window := [2, Sentence]; POSContext := NN$\mid$VB} will extract only those tokens within current sentence, which are present within a distance of 2 on both sides from the current SU and have POS tag either NN (noun singular) or VB (verb, base form).

\begin{figure*}[ht]
\begin{center}
\centerline{\includegraphics[width=\columnwidth]{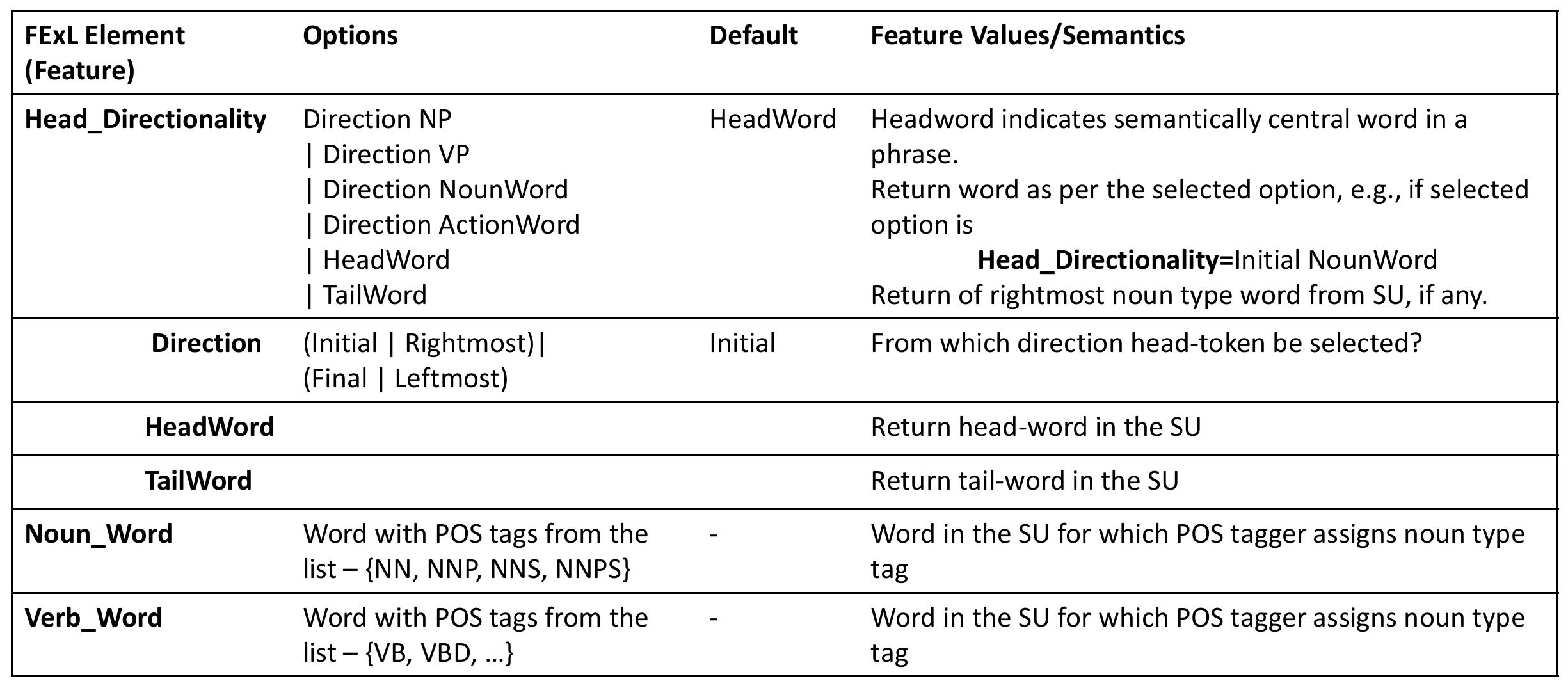}}
\caption{Specifying Head-directionality as a Feature. NP stands for Noun-Phrase and VP stands for Verb-Phrase.}
\label{head}
\end{center}
\end{figure*}

Table~\ref{head} illustrates how to specify the way head directionality of current SU should be extracted. 

\subsubsection{Semantic Similarity and Relatedness based Features}
Semantic similarity can be estimated between words, between phrases, between sentences, and between documents in a corpus. Estimation could either be based upon corpus text alone by applying approaches like vector space modeling~\cite{vsm}, latent semantic analysis~\cite{27}, topic modeling~\cite{blei2009topic}, or neural embeddings (e.g., Word2Vec~\cite{w2v} or Glove~\cite{glove}) and their extensions to phrase, sentence, and document levels. Otherwise it can be estimated based upon ontological relationships (e.g., WordNet based~\cite{4}) among concept terms appearing in the corpus. 

\subsubsection{Statistical Features}
\begin{figure*}[ht]
\begin{center}
\centerline{\includegraphics[width=\textwidth]{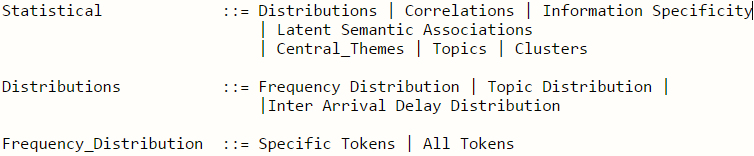}}
\caption{Statistical Features together with methods to extract these features from text at different levels}
\label{Fig9}
\end{center}
\end{figure*}

Figure~\ref{Fig9} depicts different types of statistical features which can be extracted for individual documents or corpus of documents together with methods to extract these features at different levels.  
In particular, examples of distributions which can be estimated include frequency distributions for terms, term distributions in topics and topic distributions within documents, and distribution of term inter-arrival delay, where inter arrival delay for a term measures number of terms occurring between two successive occurrences of a term.

\section{Illustration of nlpFSpL Specification}\label{nlpFSpL}

Let us consider a problem of identifying medical procedures being referenced in a medical report. \\
\textbf{Sample Input (Discharge summary):} ``“This XYZ non-interventional study report from a medical professional. Spontaneous report from a physician on 01-Oct-1900.  Patient condition is worsening day by day. Unknown if it was started before or after levetiracetam initiation. The patient had a history of narcolepsy as well as cataplexy. Patient condition has not recovered. On an unknown date patient was diagnosed to have epilepsy.  The patient received the first dose of levetiracetam for seizure."

Table~\ref{Fig15} shows specification of features in nlpFSpL.
\begin{table*}[htbp]
	\centering
	\begin{adjustbox}{width=\textwidth}
	\begin{tabular}{ll|l}
		\toprule
		\textbf{nlpFSpL Element } & \textbf{Parameter Values} & \textbf{Details} \\
		\midrule
		Syntactic\_Unit & Word & Linguistic features should be extracted for each word\\ \hline
		POS\_Sequence   & YES & POS Tag from Penn Tree Bank for each SU\\ \hline
		POS\_Regex      & NN & TRUE if POS tag of SU matches regular expression NN \\ \hline
		Suffix\_Prefix  & [Suffix, 3, NULL] & Suffix of 3 characters from SU. \\ & & NULL specifies that extracted suffix need not match any pattern. \\ \hline
		Capitalization  & First & TRUE if first character od SU is capitalized \\ \hline
		Special\_Chars  & @,- & TRUE if SU contains characters '@','-'\\ \hline
		Context\_Window & [3, Para] & Include all SUs falling within 3 distance (either side) from  current \\ & & SU. Second parameter `Para' specifies that SUs to be included \\ & &  in the context window must be present within the same \textit{Paragraph} \\ & & /though may span across sentences.
		\\
		\bottomrule
	\end{tabular}%
    \end{adjustbox}
    \caption{nlpFSpL Specification for Example 1}
	\label{Fig15}%
\end{table*}%

Table~\ref{Fig14a} and Table~\ref{Fig14b} contain corresponding feature matrix (limited to first sentence and does not involve any preprocessing of input text).

\begin{table*}[htbp]
	\centering
	\begin{adjustbox}{width=\textwidth}
	\begin{tabular}{cccccc}
		\toprule
		\textbf{Syntactic\_Unit} & \textbf{POS\_Sequence} & \textbf{POS\_Regex} & \textbf{Suffix\_Prefix} & \textbf{Capitalization} & \textbf{Special\_Chars} \\
		\midrule
		This  & DT    & FALSE  & his   & TRUE  & FALSE, FALSE \\
		XYZ   & JJ    & FALSE & XYZ   & TRUE  & FALSE, FALSE \\
		non   & AFX   & FALSE & non   & FALSE & FALSE, FALSE \\
		-     & HYPH  & FALSE & -     & FALSE & FALSE, TRUE \\
		interventional & JJ    & FALSE & nal   & FALSE & FALSE, FALSE \\
		study & NN    & TRUE & udy   & FALSE & FALSE, FALSE \\
		report & NN    & TRUE & ort   & FALSE & FALSE, FALSE \\
		from  & IN    & FALSE & rom   & FALSE & FALSE, FALSE \\
		a     & DT    & FALSE  & a     & FALSE & FALSE, FALSE \\
		medical & JJ    & FALSE & cal   & FALSE & FALSE, FALSE \\
		professional & NN    & TRUE & nal   & FALSE & FALSE, FALSE \\
		.     & .     & FALSE & .     & FALSE & FALSE, FALSE \\
		\bottomrule
	\end{tabular}%
   \end{adjustbox}	 
    \caption{Part I: Output Feature Matrix from FExSys based upon nlpFSpL Specification in Table~\ref{Fig15}}
	\label{Fig14a}%
\end{table*}%
\begin{table*}[htbp]
	\centering
	\begin{tabular}{cc}
		\toprule
		\textbf{Syntactic\_Unit} & \textbf{Context\_Window} \\
		\midrule
		This  &  XYZ, non, - \\
		XYZ   & This, non, -, interventional \\
		non   & This, XYZ, -, interventional, study \\
		-     & This, XYZ, non, interventional, study, report \\
		interventional & XYZ, non, -, study, report, from \\
		study & non, -, interventional, report, from, a \\
		report & interventional, study, from, a, medical \\
		from  & interventional, study, report, a, medical, professional \\
		a     & study, report, from, medical, professional, . \\
		medical & report, from, a, professional, ., Spontaneous \\
		professional & from, a, medical, ., Spontaneous, physician \\
		.     & a, medical, professional, Spontaneous, physician, on \\
		\bottomrule
	\end{tabular}%
	 \caption{Part II: Output Feature Matrix from FExSys based upon nlpFSpL Specification in Table~\ref{Fig15} (Note: Col 1 is repeated from Table~\ref{Fig14a} for reading convenience)}
	\label{Fig14b}%
\end{table*}%

\section{NLP Feature Reuse across TA Applications}\label{fr}
Next let us consider a case for enabling automated reuse of feature specifications in nlpFSpL across different semantically related applications. 

\subsection{Illustrative Example}\label{events}
To illustrate that semantically different yet related applications may have significant potential for reuse of features, let us consider the problem of \textit{event extraction}, which involves identifying occurrences of specific type of events or activities from raw text. 

Towards that, we analysed published works on three different types of events in different domains as described next:
\begin{description}
	\item[Bio-molecular Interactions] Objective of this study is to design a ML model for identifying if there exist mentions of one of the nine types of bio-molecular interactions in (publicly available) Biomedical data. To train SVM based classifier, authors use GENETAG database, which is a tagged corpus for gene/protein named entity recognition. BioNLP 2009 shared task test-set was used to estimate performance of the system. Further details can be found at~\cite{e1}.
	\item[Financial Events in News] Objective of the study was to design ML model for enabling automated detection of specific financial events in the news text. Ten diﬀerent types of ﬁnancial events were considered including announcements regarding CEOs, presidents, products, competitors, partners, subsidiaries, share values, revenues, proﬁts, and losses. To train and test SVM and CRF based ML models, authors used data set consisting of 200 news messages extracted from the Yahoo! Business and Technology newsfeeds, having financial events and relations manually annotated by 3 domain experts. Further details can be found at~\cite{e2}.
	\item[Events from Twitter Data] Objective of the study was to design an ML based system for extracting open domain calendar of significant events from Twitter-data. 38 different types of events were considered for designing the system. To train the ML model, an annotated corpus of 1000 tweets (containing 19,484 tokens) was used and trained model was tested on 100 million most recent tweets. Further details can be found at~\cite{e3}.
\end{description}
Table~\ref{Fig20} below depicts classes of features selected by authors of these works (as described in the corresponding references above) to highlight the point that despite domain differences, these applications share similar sets of features. Since authors of these works did not cite each other, it is possible that that these features might have been identified independently. This, in turn, supports the hypothesis that if adequate details of any one or two of these applications are fed to a system described in this work, which is designed to estimate semantic similarities across applications, system can automatically suggest potential features for consideration for the remaining applications to start with without requiring manual knowledge of the semantically related applications.   
\begin{table*}[htbp]
	\centering
	\begin{adjustbox}{width=\textwidth}
		\begin{tabular}{cccc}
			\toprule
			\textbf{Features} & \textbf{Bio-molecular Interactions} & \textbf{Financial Events in News} & \textbf{Events from Twitter Data} \\
			\midrule
			N-gram  &  \checkmark   & -  & -  \\
			POS   & \checkmark    & \checkmark & \checkmark   \\
			Context based Features   & -   & - & \checkmark\\
			Morphological Features & \checkmark  & \checkmark & \checkmark \\
			Orthographic & \checkmark   & \checkmark & \checkmark \\
			Dependency & \checkmark   & \checkmark & \checkmark \\
			Lexical & \checkmark    & \checkmark & \checkmark \\
			\bottomrule
		\end{tabular}%
	\end{adjustbox}	 
	\caption{Illustration of similarity of features across related applications in different domains}
	\label{Fig20}%
\end{table*}%

\subsection{Reuse Process}
\begin{figure*}[ht]
	\begin{center}
		\centerline{\includegraphics[scale=0.7]{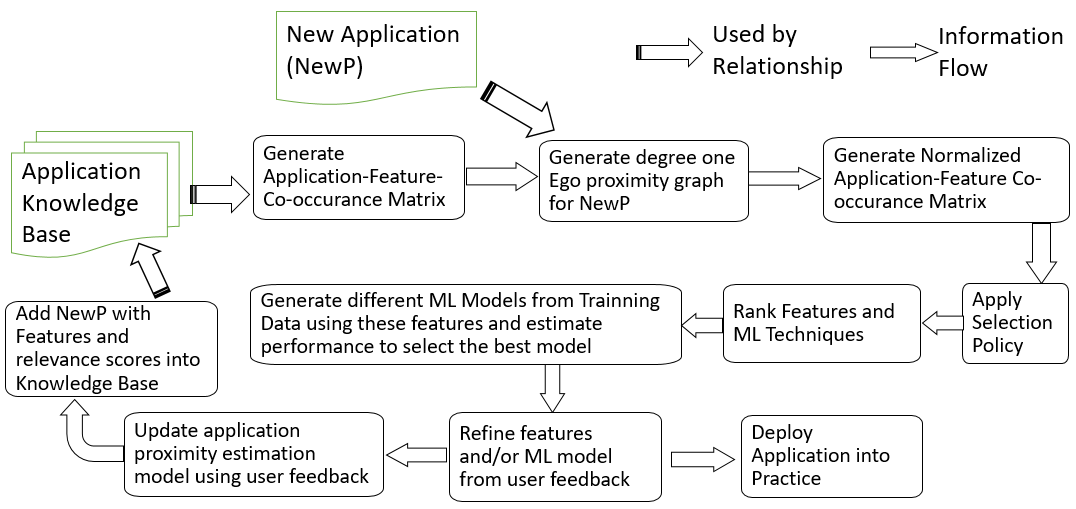}}
		\caption{High level process view for enabling automated transfer of features across semantically related applications}
		\label{Fig12}
	\end{center}
	\vskip -0.2in
\end{figure*}

Figure~\ref{Fig12} depicts overall process flow for enabling automated feature recommendations.   
For a new text analytics application requiring feature engineering, it starts with estimating its semantic proximity (from the perspective of a NLP data scientist) with existing applications with known features. Based upon these proximity estimates as well as expected relevance of features for existing applications, system would recommend features for the new application in a ranked order. Furthermore, if user's selections are not aligned with system's recommendations, system gradually adapts its recommendation so that eventually it can achieve alignment with user preferences.

Towards that let us start with characterizing text analytics applications. A TA application's details should include following fields:
\begin{description}
	\item[Problem Description] Text based description of an TA application (or problem). For example, ``identify medical procedures being referenced in a discharge summary'' or ``what are the input and output entities mentioned in a software requirements specification”.
  \item[Annotation Level] Analysis unit at which features are to be specified and training annotations are available, and ML model is designed to give outcomes. Options include word, phrase, sentence, paragraph, or document.
   \item[Problem Type] Specifies technical classification of the underlying ML challenge with respect to a well-defined ontology. E.g., Classification (with details), Clustering, etc. 
		\item[Performance Metric] Specifies how performance of the ML model is to be measured - again should be specified as per some well defined ontology. 
\end{description}

\textit{Knowledge base} of text analytics applications contains details for text analytics applications in above specified format. Each application is further assumed to be associated with a set of Features (or feature types) specified in nlpFSpL together with their relevance scores against a performance metric.  \textit{Relevance score} of a feature is a measure of the extent to which this feature contributes to achieve overall performance of ML model while solving the underlying application. Relevance score may be estimated using any of the known feature selection metrics~\cite{featureSelection}. 

To specify knowledge base formally, let us assume that there are $m$ different applications and $k$ unique feature specifications across these applications applying same performance metric. Let us denote these as follows:  $APPS=\{App_1,\ldots, App_m\}$ and ${\mathit{\Theta}}_F$ = $\left\{F_1,F_2,\dots ,F_k\right\}$ respectively.  
Knowledge base is then represented as a feature co-occurrence matrix $PF_{m\times k}$ such that $PF[i,j] = \delta_{i,F_j}$ is the relevance score of $j^{th}$ feature specification ($F_j\in \mathit{\Theta}_F$)  for $i^{th}$ application $App_i\in APPS$. 

\subsection{Measuring Proximity between Applications}
To begin, for each text based field in each TA application, pre-process text and perform term normalization (i.e., replacing all equivalent terms with one representative term in the whole corpus) including stemming, short-form and long-form (e.g., ‘IP’ and ‘Intellectual Property’), language thesaurus based synonyms (e.g., WordNet based `goal' and `objective'). 

\subsubsection{Identify key Terms}\label{ei}
Thereafter, we identify potential `entity-terms' as `noun-phrases' and `action terms' as `verb-phrases' by applying POS-Tagging and Chunking. E.g., In sentence – ``This XYZ non-interventional study report is prepared by a medical professional'', identifiable entity terms are ``this XYZ non-interventional study report'' and ``medical professional'' and identifiable functionality is `prepare'. 

\subsubsection{Generate Distributed Representations}\label{dr}
Analyze the corpus of all unique words generated from the text based details across all applications in the knowledge base. 
	Generally corpus of such textual details would be relatively small, therefore, one can potentially apply pre-trained word embeddings (e.g., word2vec~\cite{w2v} or Glove~\cite{glove}). Let $v(w)$ be the neural embedding of word $w$ in the corpus. We need to follow additional steps to generate term-level embeddings (alternate solutions also exist~\cite{phraseEmbeddings}): Represent corpus into Salton's vector space model~\cite{vsm} and estimate information theoretic weighing for each word using BM25~\cite{bm25} scheme: Let $BM25(w)$ be the weight for word $w$. Next update word embedding as $v(w)\leftarrow BM25(w)\times v(w)$. For each multi-word term $z=w_1\dots w_n$, generate term embedding by averaging embeddings of constituent words: $v(z)\leftarrow \Sigma_{i=1}^{i=n}v(w_i)$.

In terms of these embeddings of terms, for each text based field of each application in the knowledge base, generate field level embedding as a triplet as follows: 
Let $f$ be a field of an application in $APPS$. Let the lists of entity-terms and action-terms in $f$ be $en(f)$ and $act(f)$ respectively. Let remaining words in $f$ be: $r(f)$. Estimate embedding for $f$ as: $v(f)$=$[v(en(f))$, $v(act(f))$, $v(r(f))]$, where $v(en(f))$=$\Sigma_{z\in en(f)} v(z)$, $v(act(f))$=$\Sigma_{z\in act(f)}v(z)$, and $v(r(f))$=$\Sigma_{z\in r(f)}v(z)$. 
\subsubsection{Estimating Proximity between Applications}\label{ep}
After representing different fields of an application into embedding space (except AU), estimate field level similarity between two applications as follows: Let $[X_i^{en}$, $X^{act}_i$, $X_i^{r}]$ and $[X_j^{en}$, $X^{act}_j$, $X_j^{r}]$ be the representations for field $f$ for two applications $App_i$, $App_j$ $\in APPS$. In terms of these, field level similarity is estimated as $\Delta_{f}(App_i,App_j)$ = $[\Delta_{en}({f_{i}, f_j})$, $\Delta_{act}({f_{i}, f_j})$, $\Delta_{r}({f_{i}, f_j})]$, where $\Delta_{en}({f_{i}, f_j})$ = 0 \textit{if} field level details of either of the applications is unavailable \textit{else} $\Delta_{en}({f_{i}, f_j})$ = $cosine(X_i^{en}$, $X_j^{en})$; etc. 

For the field - AU, estimate $\Delta(au_i,au_j)$ = $1 \textit{ if analysis units for both applications are same i.e., } au_i = au_j$ else $0$.

In terms of these, let $\Delta(App_i,App_j)$ = $[\Delta_{en}({bd_{i}, bd_j})$, $\Delta_{act}({bd_{i}, bd_j})$, $\Delta_{r}({bd_{i}, bd_j}),$ $\Delta_{en}({dd_{i}, f_j})$, $\ldots$, $\Delta({au_{i}, au_j})]$
 be overall similarity vector across fields, where $bd$ refers to the field `problem description' etc. Finally, estimate mean similarity across constituent fields as a proximity between corresponding applications. 

\subsection{Feature Recommendations}
Let $NewP$ be new application for which features need to be specified in nlpFSpL. Represent fields of $NewP$ similar to existing applications in the knowledge base (as described earlier in the Section~\ref{dr}). 

Next, create a degree-1 ego-similarity network for $NewP$ to represent how close is $NewP$ with existing applications in $APPS$. Let this be represented as a diagonal matrix $\Delta_{m\times m}$ such that $\Delta[r,r] = \alpha_i =$ proximity between $NewP$ and $i^{th}$ application in the knowledge base (by applying steps in the Section~\ref{ep}). 

Thereafter, let $NorSim_{m\times k}=\Delta_{m\times m} \times PF_{m \times k}$ such that $NorSim[i,j]= \alpha_i \delta_{i,F_j}$ measures probable relevance of feature $F_j$ for $NewP$ w.r.t. performance metric $M$ based upon its relevance for $App_i \in APPS$. When there are multiple applications in $APPS$, we need to define a policy to determine \textit{collective probable relevance} of a feature specification in $\mathit{\Theta}_F$ for $NewP$ based upon its probable relevance scores with respect to different applications. 

To achieve that, let $Relevance$ $(NewP, f_j)$ be the relevance of $f_j$ for $NewP$ based upon a policy, which can be estimated in different ways including following:

Next, consider there different example policies: 

\begin{description}
	\item[Aggressive Policy] Weakest relevance across applications is considered:
	$Relevance(NewP,F_j)  =  \min_{i\in 1..m}{NorSim[i,j]}$
	\item[Conservative Policy] Strongest relevance across applications is considered:
	$Relevance(NewP,F_j)  =  \max_{i\in 1..m}{NorSim[i,j]}$
	\item[Probable Policy] Most likely relevance across applications is considered:
	$Relevance(NewP,F_j)  =  \frac{1}{m}\Sigma_{i\in 1..m}{NorSim[i,j]}$
\end{description} 

Rank feature specifications in $\mathit{\Theta}_F$ decreasing order based upon $Relevance(NewP,.)$, which are suggested to the NLP Data Scientist together with the supporting evidence.

\subsection{Continuous Learning from User Interactions}

There are two different modes in which user may provide feedback to the system with respect to recommended features: one where it ranks features differently and second where user provides different relevance scores (e.g., based upon alternate design or by applying feature selection techniques). Aim is to use these feed-backs to learn an updated similarity scoring function $\Delta_{new}:APPS \times APPS$ $\to$ $[0,1]$. 

In relation to $NewP$, let $Rank:{\mathrm{\Theta }}_F\times \left\{system,user\right\}\to \{1,\dots ,k\}~$ return rank of a feature and $Rel:{\mathrm{\Theta }}_F\times \left\{system,user\right\}\to [0,1]$ return relevance score of a feature based upon type -- `system' or `user'.

Next, for each $App\in APPS$, let ${Ch}\left[App\right]$ $\leftarrow$ $\emptyset$ be a hash table with keys as application ids and values as list of numbers estimated next. Also let $NewSim_{FE}\left[.\right]\leftarrow 0$ contain updated similarity scores between $NewP$ and existing applications in $APPS$.

For each feature specification $f\in \Theta_F$, determine whether `user' given rank is different from `system' given rank, i.e., $Rank(f,`system') \neq Rank(f,`user')$. If so, execute steps next.  

Let $\mathit{Bind}(f_j) \subseteq APPS$ be the list of applications, which contributed in estimating collective relevance for feature $f_j \in \Theta_F$. For example, when aggressive or conservative policy is considered, $\mathit{Bind}(f_j)$ = $\{App_r \mid Relevance(NewP,f_j)$ = $\mathit{NorSim[r,j]}\}$.

For each $App_i\in Bind\left(f_j\right)$: Add $x$ to $Ch[App_i]$, where $x$ is estimated as follows: If user provides explicit relevance scores for $App_i$, \[x = \frac{Rel(f_j, `user')}{Rel(f_j, `system')}\] 

Otherwise if user re-ranks features \[x = \frac{Rel(f_{k},`system')}{\delta_{i,f_j}}; k=Rank(f_j,`user')\] 

For each $App_i\in Bind\left(f_j\right)$: $NewSim_{FE}\left[App_i\right]\leftarrow Average(Ch[App_i])$. If $|NewSim_{FE}[App_i]$-${\alpha}_i|$ $\ge \epsilon {\alpha }_i$ i.e., when the difference between old and new similarity scores is more than $\epsilon $ fraction of original similarity, add $(\Delta(NewP,App_i),NewSim_{FE}[App_i])$ to training set $Tr_{rpls}$ so that it used to train a regression model for $\Delta_{new}(.,.)$ by applying \textit{partial recursive PLS}~\cite{qin1998recursive} with $\Delta(NewP,App_i)$ as set of predictor or independent variables and $NewSim_{FE}[App_i]$ as response variable. Existing proximity scores between applications in $APPS$ (ref. Section~\ref{ep}) are also added to training set $Tr_{rpls}$ before generating the regression model.  

Note that $\epsilon $ is a small fraction $>$ 0 which controls when should similarity model be retrained. For example, $\epsilon = 0.05$ would imply that if change in similarity is more than 5\% only then it underlying model should use this feedback for retraining.

\section{Conclusion}
In this paper, we have presented high level overview of a feature specification language for ML based TA applications and an approach to enable reuse of feature specifications across semantically related applications. Currently, there is no generic method or approach, which can be applied during TA applications' design process to define and extract features for any arbitrary application in an automated or semi-automated manner primarily because there is no standard way to specify wide range of features which can be extracted and used. We considered different classes of features including linguistic, semantic, and statistical for various levels of analysis including words, phrases, sentences, paragraphs, documents, and corpus. As a next step, we presented an approach for building a recommendation system for enabling automated reuse of features for new application scenarios which improves its underlying similarity model based upon user feedback.

To take this work forward, it is essential to have it integrated to a ML platform, which is being used by large user base for building TA applications so that to be able to populate a repository of statistically significant number of TA applications with details as specified in Section~\ref{fr} and thereafter refine the proposed approach so that eventually it rightly enables reuse of features across related applications.  

\bibliographystyle{unsrt}

\end{document}